\def\paperID{41100}
\def\confName{CVPR}
\def\confYear{2026}

\def\paperTitle{Pantheon360: Taming Digital Twin Generation via 3D-Aware 360° Video Diffusion}

\def\authorBlock{
    Ting-Hsuan Chen$^{1}$\footnotemark[1]
    \quad
    Ying-Huan Chen$^{2}$\addtocounter{footnote}{-1}\addtocounter{Hfootnote}{-1}\footnotemark[1]
    \quad
    Tao Tu$^{3}$
    \quad
    Jie-Ying Lee$^{2}$
    \quad
    Cho-Ying Wu$^{4}$
    \\
    Fangzhou Lin$^{4}$
    \quad
    Hengyuan Zhang$^{4}$
    \quad
    David Paz$^{4}$
    \quad
    Xinyu Huang$^{4}$
    \\
    Yuliang Guo$^{4}$\footnotemark[2]
    \quad
     Yu-Lun Liu$^{2}$\footnotemark[2]
    \quad
     Yue Wang$^{1}$\footnotemark[2]
    \quad
     Liu Ren$^{4}$\vspace{0.5em}
    \\
    \centerline{$^1$University of Southern California \quad $^2$National Yang Ming Chiao Tung University}
    \\
    \centerline{$^3$Cornell University \quad $^4$Bosch Research}
}

\newif\ifreview 
\newif\ifarxiv \newcommand{\arxiv}{\arxivtrue}
\newif\ifcamera 
\newif\ifrebuttal 

\arxiv 

\pdfoutput=1
\documentclass[10pt,twocolumn,letterpaper]{article}
\ifreview \usepackage[review]{cvpr} \fi
\ifarxiv \usepackage[pagenumbers]{cvpr} \fi
\ifrebuttal \usepackage[rebuttal]{cvpr} \fi
\ifcamera \usepackage{cvpr} \fi

\usepackage{graphicx}	
\usepackage{amsmath}	
\usepackage{amssymb}	
\usepackage{booktabs}
\usepackage{times}
\usepackage{microtype}
\usepackage{epsfig}
\usepackage{caption}
\usepackage{float}
\usepackage{placeins}
\usepackage{color, colortbl}
\usepackage{stfloats}
\usepackage{enumitem}
\usepackage{tabularx}
\usepackage{xstring}
\usepackage{multirow}
\usepackage{xspace}
\usepackage{url}
\usepackage{subcaption}
\usepackage{xcolor}
\usepackage[hang,flushmargin]{footmisc}

\ifcamera \usepackage[accsupp]{axessibility} \fi

% ...

\ifarxiv  \fi

\newcommand{\R}[1]{{%
    \textbf{%
        \ifstrequal{#1}{1}{\textcolor{red}{R#1}}{%
        \ifstrequal{#1}{2}{\textcolor{blue}{R#1}}{%
        \ifstrequal{#1}{3}{\textcolor{magenta}{R#1}}{%
        \ifstrequal{#1}{4}{\textcolor{teal}{R#1}}{%
                           \textcolor{cyan}{R#1}%
        }}}}%
    }%
}}

\makeatletter
\renewcommand{\paragraph}{%
    \@startsection{paragraph}{4}%
    {\z@}{-0.5em}{-0.5em}%
    {\normalfont\normalsize\bfseries}%
}
\makeatother  

\usepackage{xr-hyper}

\makeatletter
\newcommand*{\addFileDependency}[1]{
  \typeout{(#1)}
  \@addtofilelist{#1}
  \IfFileExists{#1}{}{\typeout{No file #1.}}
}

\makeatother
\newcommand*{\myexternaldocument}[1]{
    \externaldocument{#1}
    \addFileDependency{#1.tex}
    \addFileDependency{#1.aux}
}

\definecolor{cvprblue}{rgb}{0.21,0.49,0.74}
\usepackage[pagebackref,breaklinks,colorlinks,allcolors=cvprblue]{hyperref}
\usepackage[capitalize]{cleveref}
\crefname{section}{Sec.}{Secs.}
\crefname{table}{Table}{Tables}
\crefname{figure}{Fig.}{Figs.}

\ifarxiv \crefname{appendix}{App.}{Apps.}
\else \crefname{appendix}{Suppl.}{Suppls.} \fi

\unless\ifarxiv \myexternaldocument{_supplementary} \fi

\begin{document}
\pagestyle{empty}
\title{\paperTitle}
\author{\authorBlock}

\twocolumn[{
\renewcommand\twocolumn[1][]{#1}
\maketitle
\begin{center}
    \captionsetup{type=figure}
    \includegraphics[width=\textwidth]{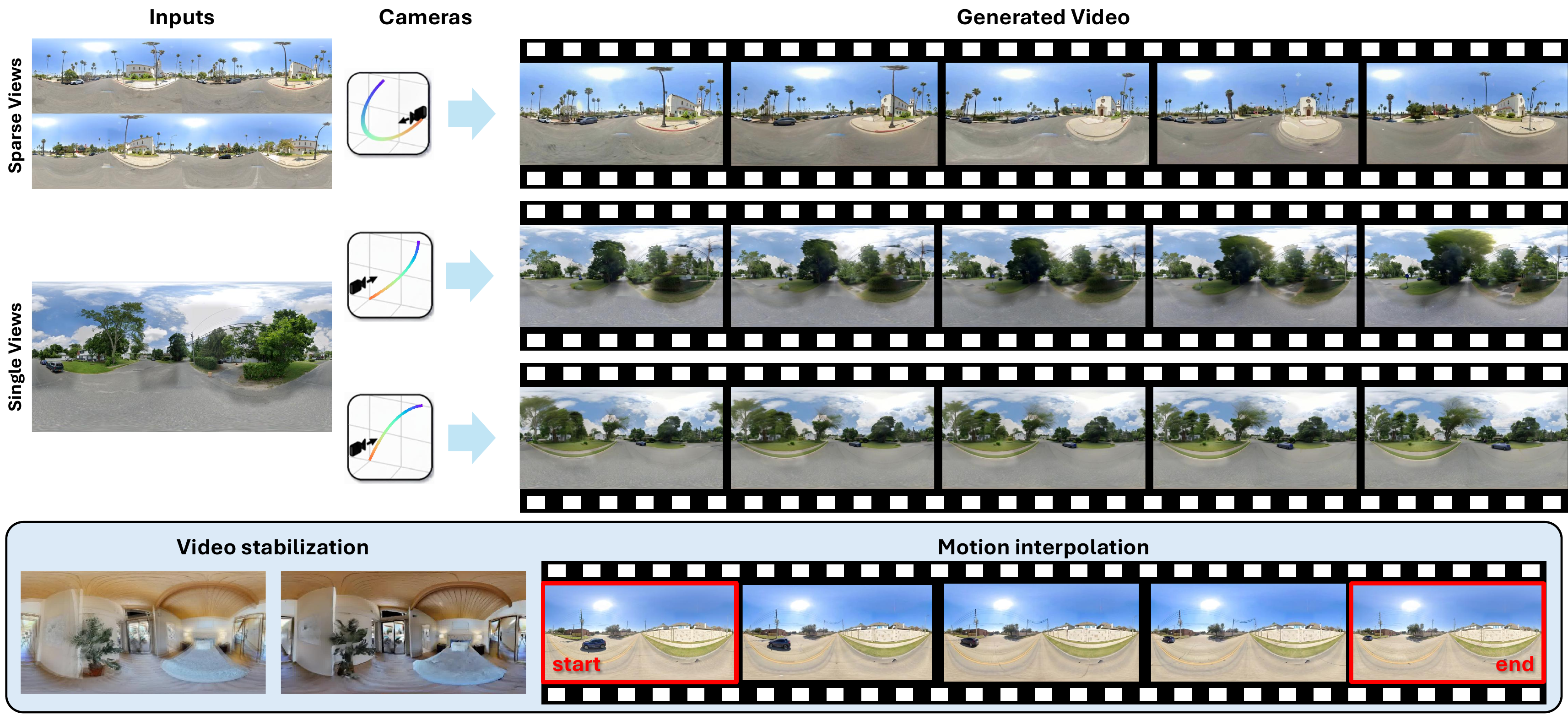}
    \vspace{-6mm}
    \captionof{figure}{
    \textbf{Pantheon360: Controllable 360° Video Generation.} Given sparse or single 360° input images, Pantheon360 generates temporally consistent 360° videos along user-defined camera trajectories with precise geometric control. \textit{Top:} From sparse views or a single view, our method synthesizes smooth videos following diverse camera trajectories across varied scenes, demonstrating flexible trajectory control from minimal input. \textit{Bottom:} Our framework enables practical applications, including video stabilization (left, transforming shaky footage into smooth output) and motion interpolation (right, generating smooth transitions between distant anchor frames marked in red).
    }
    \label{fig:teaser}
\end{center}
}]

\renewcommand{\thefootnote}{\fnsymbol{footnote}}
\footnotetext[1]{The first two authors contributed equally to this work.}
\footnotetext[2]{Equal advising.}

\maketitle

\begin{abstract}
Generating complete digital twins from videos requires precise camera control, global scene coverage, and strict spatial–temporal consistency constraints that remain challenging for perspective video generators due to their limited field of view (FoV). Their narrow FoV forces long or multi-view trajectories, amplifying cross-view inconsistency and temporal drift.
We argue that 360° video generation offers a natural solution: panoramic coverage simplifies trajectory design and provides a strong global context for maintaining coherence. We introduce \paperTitle, a controllable 360° video generation framework that synthesizes high-fidelity videos from sparse 360° inputs. The key idea is an explicit 3D Cache, reconstructed from the input, which serves as a geometric scaffold for any user-defined camera path. This allows the diffusion model to focus on photorealistic texture refinement while the 3D Cache enforces global geometric consistency.
Experiments show that Pantheon360 achieves superior visual quality and unmatched geometric coherence, enabling reliable and flexible 360° scene generation for downstream simulation and digital-twin applications.
Project page: \url{https://koi953215.github.io/pantheon360_page/}
\end{abstract}
\section{Introduction}

The creation of dynamic, complete digital twins is a fundamental goal for next-generation simulation, enabling complex, closed-loop evaluation and training for robotics and autonomous agents~\cite{cornelisse2024human,jia2024bench,pmlr-v87-kalashnikov18a,rusnak2025videoartgs,yang2025building}. Traditional 3D reconstruction can capture static scenes~\cite{dong2025digital,mohiuddin2024efficient, lin2025longsplat, mildenhall2021nerf, kerbl20233d}, but generative video models promise a revolutionary alternative: creating dynamic, photorealistic worlds with far less human effort~\cite{blattmann2023stable,brooks2024video, agarwal2025cosmos, liu2023robust}. However, this shift to generation poses new, difficult challenges, particularly in achieving 3D-aware controllability and long-term temporal consistency~\cite{he2024cameractrl,wang2024motionctrl}.

\begin{figure*}[bt!]
    \centering
    \vspace{-3mm}
    \includegraphics[width=1.0\linewidth]{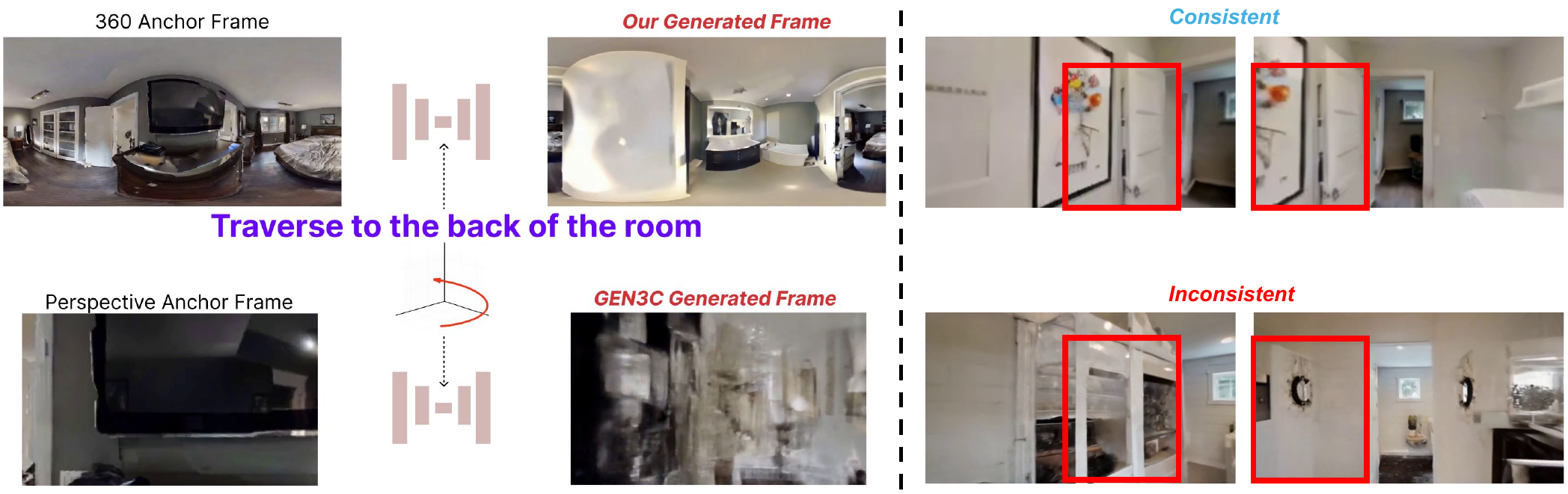}
    \vspace{-6mm}
    \caption{\textbf{Motivation for Using 360° Images for Generation.} \textit{Left:} When traversing to the back of the room, 360° anchor frames provide complete scene context, enabling accurate generation of occluded regions. In contrast, perspective anchor frames have a limited field-of-view and must hallucinate unseen areas, leading to significant artifacts. \textit{Right:} Generating 360° outputs in a single pass ensures global coherence and cross-view consistency. Our method maintains consistent object structures (red boxes highlight the same door/cabinet viewed from different angles), while GEN3C's perspective-based generation produces geometrically inconsistent results across views.}
    \label{fig:360_pers}
\end{figure*}

The dominant paradigm, camera-controlled perspective video generation, is fundamentally unsuitable for this task~\cite{ma2024trailblazer,guo2024cameractrl}. It suffers from a limited field-of-view (FoV), rendering it blind to most of the scene from its initial frame. When simulating complex, long trajectories or multi-trajectory exploration, the model must repeatedly guess and hallucinate unseen regions. This leads to redundant conditioning, processing the same geometry from different views, and, inevitably, severe spatial and temporal inconsistencies as the generated world contradicts itself, as illustrated in Fig.~\ref{fig:360_pers}.

The 360° video format offers a clear solution~\cite{tan2024imagine360, tan2025argus}. By capturing the entire scene's context from $t=0$, it provides a holistic understanding that perspective models lack, simplifying trajectory representation and dramatically improving consistency. However, 360° video generation introduces its own unique challenges, namely the extreme distortion of equirectangular projection and, most critically, the difficulty of precise geometric control. Existing controllable 360° models, such as GenEX~\cite{lu2025genexgeneratingexplorableworld}, are limited to simple, high-level action control, moving forward, rather than exact camera trajectory following. Others, like CamPVG~\cite{ji2025campvg}, only validate on synthetic data, failing to address the complexity of in-the-wild scenes.

To solve this, we present Pantheon360. Our framework is enabled by recent advances in powerful 3D foundation models~\cite{thengane2025foundational,dong2023leveraging,wang2024dust3r}. We leverage these models, such as PI3 ~\cite{wang2025pi3}, VGGT ~\cite{wang2025vggt}, to establish a robust geometric prior for the scene. This leads to our core design: to assign complex 3D geometric reasoning to an explicit 3D Cache, thereby allowing the diffusion model to focus its generative power solely on photorealistic texture synthesis. We introduce this 3D Cache, a 3D point cloud representation of the scene, which is efficiently reconstructed from sparse 360° inputs at inference time.

Our generative process operationalizes this decoupling. First, we render the 3D point cloud along the exact user-defined camera trajectory $C_{target}$. This produces a geometry-only video ($V_{geo}$) that serves as a strong, 3D-consistent scaffold. Second, our fine-tuned diffusion model is conditioned on this $V_{geo}$ scaffold and semantic features from the input. In this way, global geometric consistency is strictly enforced by the 3D Cache, while the diffusion model handles the photorealistic synthesis.

The robust geometric control and 3D-aware synthesis of Pantheon360 unlock numerous downstream applications. We demonstrate state-of-the-art performance against both perspective and 360° baselines, and showcase its utility in novel 360° interpolation, for example, stitching Google Maps Street View data, and video stabilization tasks.

Our main contributions are:
\begin{itemize}
    \item We enable exact camera trajectory control for in-the-wild 360° videos, overcoming limitations of prior methods restricted to simple action control or synthetic data.
    
    \item We propose Pantheon360, a novel framework that achieves this precise control by using an explicit 3D Cache to enforce geometric consistency, allowing the diffusion model to focus solely on photorealistic texture refinement.
    
    \item We demonstrate state-of-the-art performance in 360° video synthesis across various tasks and validate its utility in downstream applications like 360° interpolation and stabilization.
\end{itemize}

\section{Related Work}

\paragraph{Camera-Controllable Video Generation.}
Achieving precise camera control is a major goal in video generation. Existing approaches can be broadly categorized into parametric and geometric methods. Parametric methods embed camera information through direct parameters (e.g., rotation matrices, translation vectors)\cite{wang2024motionctrlunifiedflexiblemotion, vanhoorick2024generativecameradollyextreme} or Plucker coordinate embeddings\cite{fu20253dtrajmastermastering3dtrajectory, wang2024cpacameraposeawarenessdiffusiontransformer,liang2025wonderlandnavigating3dscenes, he2025cameractrlenablingcameracontrol,bahmani2025ac3danalyzingimproving3d, watson2025controllingspacetimediffusion}, offering lightweight solutions. Training-free methods~\cite{hou2025trainingfreecameracontrolvideo, you2025nvssolvervideodiffusionmodel} leverage pretrained video diffusion priors to achieve camera control without additional training. Recent works have also extended camera control to dynamic scenes~\cite{mark2025trajectorycrafter, xiao2024trajectory, he2025cameractrl, wu2025cat4d, zhang2025recapture, huang2025vivid4d, zhang2025spatialcrafter, shih2025prior, fan2025spectromotion, chien2025splannequin, liu2023robust}. In contrast, geometric methods~\cite{ren2025gen3c, yu2024viewcrafter, mark2025trajectorycrafter, chen2025flexworld, liu2024reconx, yan2024streetcrafter, chen2024mvsplat360, wang2025vistadream, ma2025you, fischer2025flowr, wang2025videoscene, yin2025gsfixer, wu2025genfusion, lin2025longsplat} leverage explicit 3D representations by reconstructing the scene geometry and rendering it along the target path. This ``3D cache'' paradigm enforces 3D consistency by grounding generation in geometric structure. However, existing methods are primarily designed for planar perspective videos with limited field-of-view (FoV), constraining their ability to fully observe the complete scene. Our work extends the 3D-cache approach to the 360° domain, leveraging holistic 360° inputs to naturally overcome FoV limitations and enable comprehensive scene understanding.

\paragraph{360° Video Generation.}
Directly generating 360° video presents unique challenges, including handling equirectangular distortion~\cite{wang2024depth} and ensuring seamless panoramic continuity. Early works in this space focused on text-to-360°, image-to-360° synthesis, or scene inpainting~\cite{chen2022text2light, wang2022stylelight, akimoto2022diverse, wu2023panodiffusion, liao2023cylin, feng2023diffusion360, wang2024customizing, zhang2024taming, kalischek2025cubediff, yuan2025camfreediff, wu2023panodiffusion, tang2023mvdiffusionenablingholisticmultiview, oh2022bips, kalischek2025cubediff, wang2024360dvd, wu2025aurafusion360}. While capable of producing panoramic content, these methods generally lack mechanisms for complex or precise camera control. Other methods~\cite{tan2025argus, tan2024imagine360} address a different task of converting perspective videos to 360° panoramas.
More recent models have begun to tackle direct 360° control, but still fall short. GenEx~\cite{lu2025genexgeneratingexplorableworld} (as discussed in our experiments) is a notable 360° world model, but it focuses on high-level, action-based'' control. It can support simple actions like ``move forward'' or ``rotate,'' but cannot follow an exact, pre-defined camera trajectory. Concurrently, CamPVG~\cite{ji2025campvg} (also in our experiments) has demonstrated promise in precise trajectory following, but it is validated primarily on synthetic datasets. This leaves its applicability to diverse, in-the-wild videos with complex, real-world trajectories unproven.
In contrast, our Pantheon360 pioneers exact camera trajectory control for in-the-wild 360° videos by integrating a robust 360-aware 3D cache with a generative model trained on real-world 360° data.

\paragraph{360° Reconstruction Models.}
Reconstructing 3D scenes from 360° inputs is a related but distinct problem. Methods like ~\cite{chen2023panogrf, chen2025splatter, ren2025panosplatt3r, coors2018spherenet, garanderie2018eliminating, eder2019pano, tateno2018distortion, zioulis2018omnidepth, zhuang2022acdnet, zhang2025pansplat, lee2025skyfall, wu2025aurafusion360, fan2025spectromotion} aim to faithfully reproduce input views and interpolate between them. However, these are fundamentally reconstruction models, not generative models—they excel at novel view synthesis for seen regions but cannot creatively hallucinate plausible content for large occluded or entirely unseen areas. In contrast, our method uses 3D reconstruction only as a 3D Cache, while the final photorealistic synthesis and generative completion of unseen regions is handled by our video diffusion model trained on real-world 360° data.

\section{Method}

We introduce \textbf{Pantheon360}, a novel framework for controllable $360^{\circ}$ video synthesis from sparse inputs. Our method is built upon a pre-trained latent video diffusion model, SVD~\cite{blattmann2023stable}, but introduces a robust conditioning mechanism guided by an explicit 3D scene representation.

\begin{figure*}[bt!]
    \centering
    \vspace{-3mm}
    \includegraphics[width=1.0\linewidth]{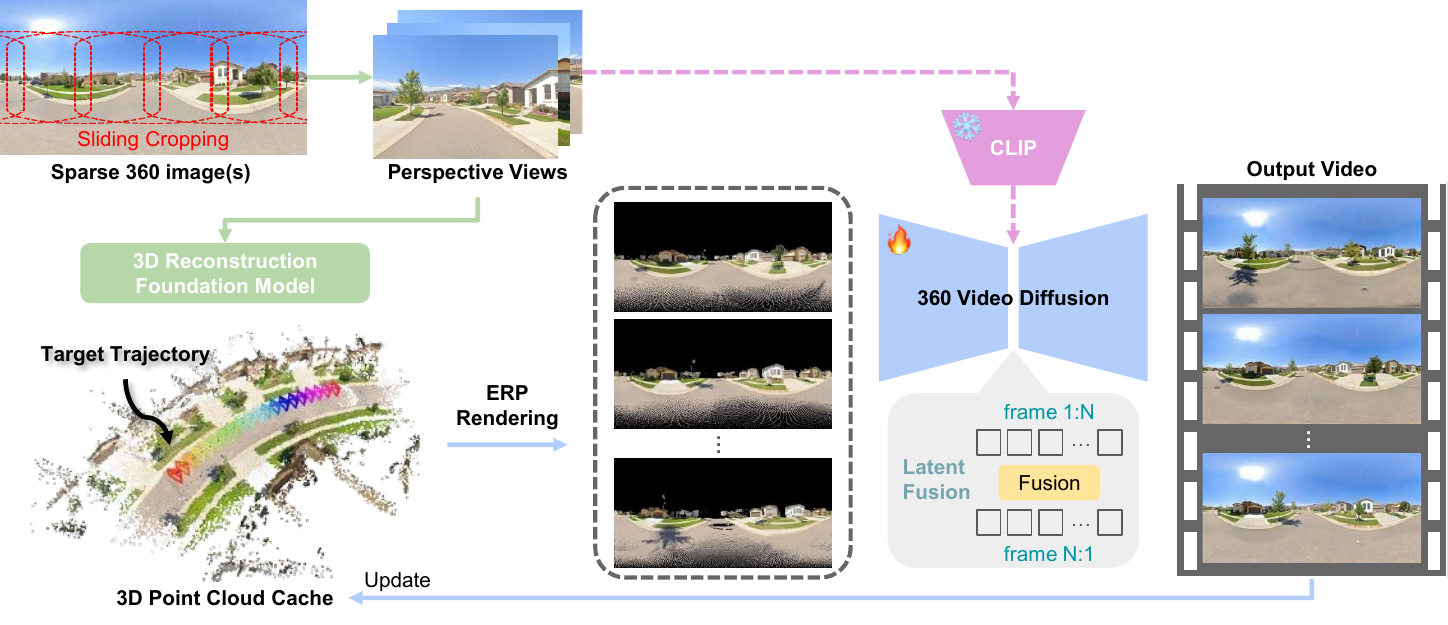}
    \vspace{-6mm}
    \caption{\textbf{Pantheon360 Pipeline.} Given sparse 360° input frames, we first crop them into perspective views and reconstruct a 3D point cloud cache using foundation models (e.g., $\pi^3$~\cite{wang2025pi3}, VGGT~\cite{wang2025vggt}). We then render this cache along the target camera trajectory to produce a geometry-only equirectangular video $V_{geo}$, which is encoded into latent space and concatenated with noised latents for geometric conditioning. Simultaneously, CLIP features extracted from 8 perspective crops of the first frame provide semantic conditioning via cross-attention. Our fine-tuned video diffusion model leverages both geometric and semantic conditions to generate temporally consistent, photorealistic 360° videos with precise trajectory control. For interpolation tasks, we employ dual-anchor latent fusion~\cite{feng2024explorative} to blend information from both start and end frames, ensuring smooth transitions between distant viewpoints.}
    \label{fig:pipeline}
\end{figure*}

\subsection{Problem Formulation}

Given sparse $360^{\circ}$ input frames $\{I_k\}$ and a target camera trajectory $C_{target} = \{c_1, \dots, c_T\}$, our goal is to generate a temporally consistent $360^{\circ}$ video $Y_{equi} \in \mathbb{R}^{T \times 3 \times H' \times W'}$ in equirectangular format. Our approach leverages two key elements: an explicit $3D$ Cache for geometric condition and $360^{\circ}$ video generation for global consistency.

\subsection{3D-Aware 360° Video Generation}
\paragraph{3D Cache Reconstruction.} 
At inference time, we first reconstruct the 3D Cache from the sparse input frames $\{I_k\}$. We crop each 360° frame into multiple perspective views and feed them into 3D reconstruction methods, such as PI3~\cite{wang2025pi3} or VGGT~\cite{wang2025vggt}, to produce a 3D point cloud that explicitly models the scene's spherical geometry. Our framework is compatible with any method that can generate this point cloud representation~\cite{wang2025continuous, wang2024dust3r, leroy2024mast3r}.

\paragraph{Geometric Conditioning ($V_{geo}$).} 
We condition our diffusion model on explicit 2D renderings from this 3D cache. Given the user-defined trajectory $C_{target}$, we render the 3D point cloud into equirectangular projection (ERP) format along this trajectory to produce a geometry-only video $V_{geo} \in \mathbb{R}^{T \times 3 \times H' \times W'}$. This $V_{geo}$ is then passed through the VAE encoder $\mathcal{E}$ to produce a latent scaffold $v_{equi} = \mathcal{E}(V_{geo})$, which is concatenated with the noised latent at each diffusion step to guide the video generation process with precise geometric information.

\subsection{Model Architecture}

Our generator $G$ is a fine-tuned SVD U-Net $f_{\theta}$. We adopt the pre-trained SVD VAE Encoder $\mathcal{E}$ and Decoder $\mathcal{D}$. The denoising U-Net $f_{\theta}$ is conditioned on two streams:

\paragraph{Geometric Latent (via concatenation).} 
The geometry-only video $V_{geo}$ is passed through the VAE encoder $\mathcal{E}$ to produce a latent scaffold $v_{equi} = \mathcal{E}(V_{geo})$. This latent is concatenated with the noised ground truth latent $y_{equi, t}$ at each diffusion step, serving as our 3D-aware geometric condition.

\paragraph{Image Features (via cross-attention).} 
To provide semantic information, we extract features from the first frame $I_0$. Since CLIP provides more robust features from perspective views than from distorted equirectangular images, we crop $I_0$ into 8 perspective frames (every 45° of yaw), pass them through CLIP extractor $\mathcal{F}$, and concatenate the resulting features to form $c_{img}$ for cross-attention conditioning.

\subsection{Model Training}
Our generative model is a 3D-aware 360° video diffusion model, adapted for equirectangular projection. Its primary objective is to synthesize photorealistic 360° video frames $Y_{equi}$ conditioned on our explicit geometric scaffold $V_{geo}$ and sparse input semantic features $C_{img}$.
We employ a standard diffusion objective to train the model $f_{\theta}$ to denoise a noisy latent representation $y_{equi, t}$ back to the ground-truth video latent $y_{equi}$:

$$
L = \mathbb{E}_{y_{equi}, v_{equi}, c_{img}, t, \epsilon} [ \lambda(t) || \epsilon - f_{\theta}(y_{equi, t}, t, v_{equi}, c_{img}) ||_2^2 ]
$$
where $y_{equi} = \mathcal{E}(Y_{equi})$ is the latent of the ground-truth video, $y_{equi, t}$ is its noised version at timestep $t$, $v_{equi} = \mathcal{E}(V_{geo})$ is the latent representation of our geometric scaffold, and $c_{img}$ represents concatenated semantic features derived from the sparse input image. This formulation explicitly injects the 3D geometric information ($v_{equi}$) and semantic context ($c_{img}$) into the denoising process, guiding the generation towards geometrically consistent and photorealistic 360° videos. The detailed process for curating our 360° dataset and generating the $(Y_{equi}, V_{geo})$ training pairs is described in Sec.~\ref{sec:data_curation}.

\paragraph{Implementation Details.} 
Both single-anchor and dual-anchor models are trained at $1024 \times 512$ resolution on 4 A100 GPUs for 5 days each. For 3D reconstruction, we use PI3~\cite{wang2025pi3} with a confidence threshold of 0.25 and sky masking. Full details are in the supplementary material.

\subsection{Data Curation and Training Data Generation}
\label{sec:data_curation}

\paragraph{Data Source and Curation.} 
Our primary goal is to generate controllable video for in-the-wild scenes, not just synthetic environments. To achieve this robustness, we leverage the \textbf{360-1M}~\cite{wallingford2024image}, a large-scale collection of diverse, real-world 360° videos. We adopt a comprehensive filtering pipeline to remove low-quality content, such as mislabeled 180° videos, static posters, and clips with low motion, using this final filtered dataset as our foundation.

\paragraph{On-the-fly Data Annotation and Generation.} 
A major challenge is that 360-1M is unlabeled; it provides raw video clips but lacks the ground-truth camera poses and 3D geometry required for our 3D-aware training. To prepare the required training pairs $(Y_{equi}, V_{geo})$, we generate these annotations on-the-fly.

For each ground-truth $360^{\circ}$ video $Y_{GT}$ sampled from the dataset, we set the ground-truth target video $Y_{equi} = Y_{GT}$. We then auto-annotate the 3D Cache and ground-truth trajectory by processing the entire video $Y_{GT}$ using ViPE~\cite{huang2025vipe}, which excels at robust 3D estimation for 360° video. We denote the estimated camera pose trajectory as $C_{GT\_poses}$ and use the resulting SLAM~\cite{teed2021droid} generated by ViPE's optimization as our 3D Cache. This step is crucial, as these SLAM points represent the most geometrically robust features in the scene. Using a high-quality, non-noisy point cloud ensures the model learns to trust the geometric condition ($V_{geo}$), rather than learning to ignore it due to poor geometry. Finally, we generate the geometric scaffold $V_{geo}$ by setting our target path $C_{target} = C_{GT\_poses}$ and rendering the high-fidelity 3D Cache along this ground-truth trajectory.

\subsection{Dual-Anchor Latent Fusion for Interpolation}

While our primary model is conditioned on a single start frame, we also train a dual-anchor variant conditioned on both start and end frames to enable precise interpolation between sparse observations. However, we observe that even the dual-anchor model can fail when the reconstructed 3D Cache quality is suboptimal. Due to sparse input views, the point cloud geometry can be inconsistent with the target end frame, leading to sudden jumps or discontinuities in the generated video. To address this issue, we adopt the latent fusion technique from Time Reversal Fusion~\cite{feng2024explorative}, which smoothly blends information from both anchor frames at the latent level, effectively mitigating these geometric inconsistencies while maintaining temporal smoothness. This technique proves especially valuable for real-world scenarios with challenging reconstruction conditions, such as Google Maps Street View synthesis. We validate the effectiveness of this approach in Sec.~\ref{sec:ablation}.
\section{Experiments}

Pantheon360 is designed to perform precise trajectory-controlled 360° video generation by leveraging an explicit 3D Cache. We validate its effectiveness through extensive experiments across multiple tasks: single 360° view-to-video generation, sparse 360° views-to-video generation. We further compare against 360° reconstruction method and 360° world models qualitatively and demonstrate practical applications.

\subsection{Single 360° View-to-Video Generation}

Pantheon360 generates video from a single 360° image by first building a 3D Cache via PI3~\cite{wang2025pi3}, rendering it along the target trajectory into a geometric scaffold $V_{geo}$, and feeding it into the video diffusion model.

\paragraph{Evaluation and Baselines.} 
We compare Pantheon360 to three baselines adapted from controllable perspective video generation: ViewCrafter~\cite{yu2024viewcrafter}, TrajectoryCrafter~\cite{mark2025trajectorycrafter}, and GEN3C~\cite{ren2025gen3c}. Since these methods are designed for perspective inputs, we adapt them to the 360° domain by rendering our geometric scaffold $V_{geo}$ (equirectangular format) and cropping it into 8 perspective views (one every 45°) as their 3D-aware condition. We evaluate all methods on the Web360 dataset~\cite{wang2024360dvd}, which contains approximately 2,000 diverse in-the-wild 360° video clips primarily in outdoor environments. We randomly sample 100 test sequences. Following prior work, we report PSNR, SSIM, LPIPS, and FVD for pixel-level quality, and MET3R~\cite{asim24met3r} for 3D geometric consistency. All metrics are computed on 8 perspective crops extracted at 45° yaw intervals from ERP outputs for fair comparison.

\begin{figure*}[h!]
    \centering
    \vspace{-3mm}
    \includegraphics[width=1.0\linewidth]{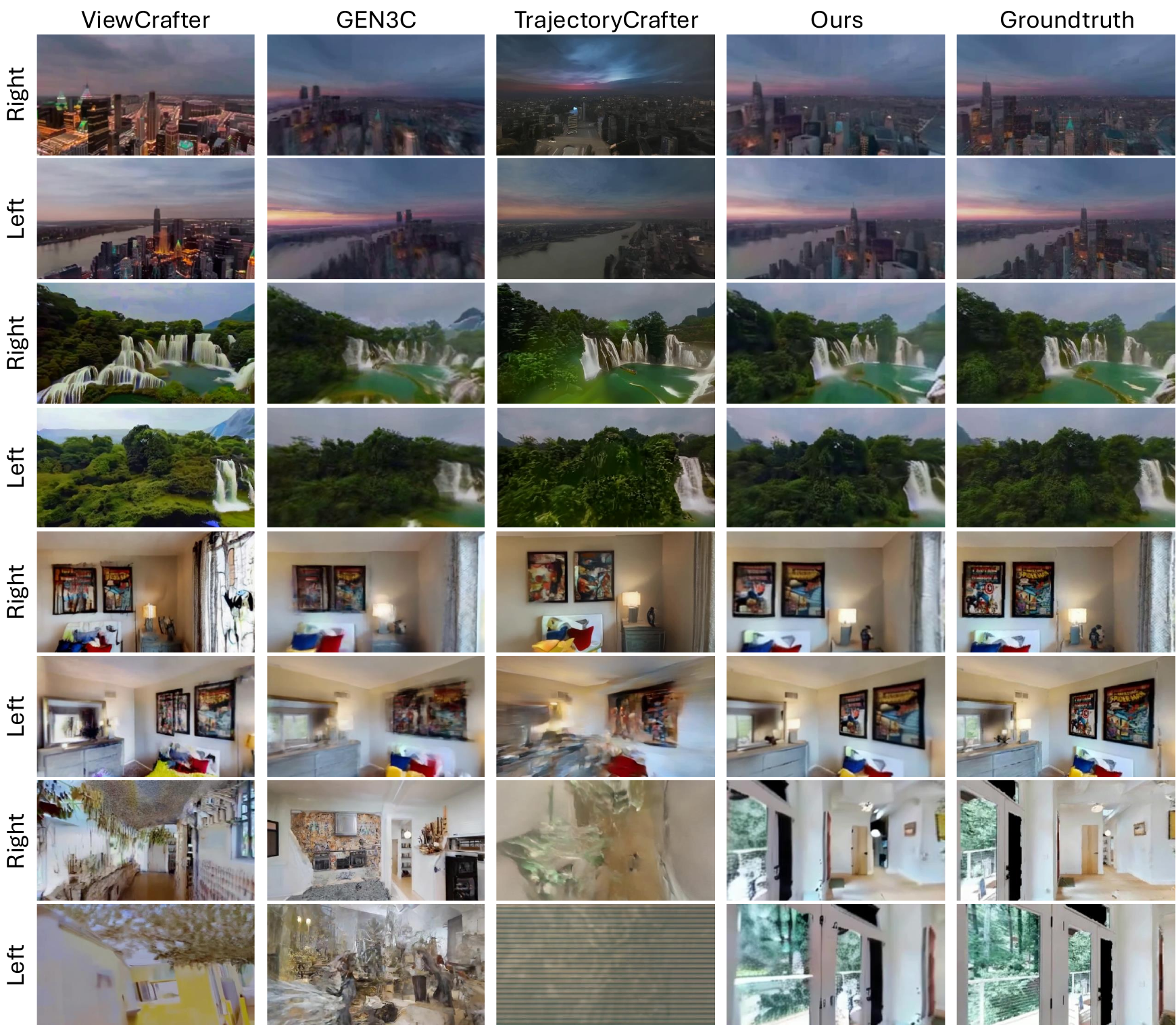}
    \vspace{-6mm}
    \caption{\textbf{Visualization Results on the Web360~\cite{wallingford2024image} and Habitat~\cite{habitat19iccv} datasets.} Our method (Ours) generates temporally consistent videos with coherent cross-view geometry across diverse camera trajectories. In contrast, perspective-based baselines (ViewCrafter~\cite{yu2024viewcrafter}, TrajectoryCrafter~\cite{mark2025trajectorycrafter}, GEN3C~\cite{ren2025gen3c}) exhibit severe cross-view inconsistencies when rendered from different viewing angles (Left vs. Right), revealing their limited ability to maintain geometric coherence across viewpoints. This inconsistency is particularly pronounced when the initial frame captures geometry at close range, where the limited field-of-view fails to provide sufficient spatial context for consistent generation. Our 360° approach naturally overcomes these limitations through complete panoramic coverage, ensuring globally consistent generation across all viewpoints.}
    \vspace{-5pt}
    \label{fig:qaulitatively}
\end{figure*}

\paragraph{Results.} 
Quantitative results are provided in Table~\ref{tab:single_view}. Pantheon360 significantly outperforms all baselines across all metrics. The superior performance stems from 360° videos' full panoramic field-of-view, which provides better cross-view consistency and enables the diffusion model to better understand the complete scene. Qualitative comparisons are shown in Fig.~\ref{fig:qaulitatively}.

\begin{table}[t]
\centering
\caption{\textbf{Quantitative comparison on single 360° view-to-video generation on Web360 dataset.} $\downarrow$ indicates lower is better, $\uparrow$ indicates higher is better.}
    \vspace{-3mm}
\label{tab:single_view}
\resizebox{\columnwidth}{!}{
\begin{tabular}{lccccc}
\toprule
\textbf{Method} & \textbf{FVD} $\downarrow$ & \textbf{SSIM} $\uparrow$ & \textbf{PSNR} $\uparrow$ & \textbf{LPIPS} $\downarrow$ & \textbf{MET3R} $\downarrow$ \\
\midrule
ViewCrafter & 525.746 & 0.371 & 15.654 & 0.284 & 0.4914 \\
TrajectoryCrafter & 517.475 & 0.454 & 15.151 & 0.219 & 0.4578 \\
GEN3C & 380.080 & 0.583 & 20.730 & 0.145 & 0.3496 \\
\midrule
Pantheon360 (Ours) & \textbf{356.151} & \textbf{0.746} & \textbf{22.838} & \textbf{0.065} & \textbf{0.2840} \\
\bottomrule
\end{tabular}
}
\end{table}

\subsection{Sparse 360° Views-to-Video Generation}

We further apply Pantheon360 to a sparse-view setting, where multiple 360° keyframes are provided at different time steps. Similar to the single-view task, we first predict the depth for each view using PI3~\cite{wang2025pi3}, create the 3D Cache from these sparse views, and use the camera trajectory to render the geometric scaffold into videos, which are fed into Pantheon360 to generate the output video.

\paragraph{Evaluation and Baselines.} 
We compare our method to the same three baselines (ViewCrafter, TrajectoryCrafter, GEN3C) using their 8-crop adaptations. We evaluate on the Habitat dataset~\cite{habitat19iccv}, which provides 34,000 synthetic 360° video clips in indoor environments using reconstructions. The trajectories are non-looped polylines with diverse and complex navigation patterns, making this dataset particularly suitable for evaluating sparse-view video generation with challenging camera motion. We randomly sample 50 test sequences with ground-truth camera poses for rigorous quantitative evaluation.

\paragraph{Results.} 
Quantitative results are provided in Table~\ref{tab:sparse_view}. Pantheon360 again achieves the best performance across all metrics, with particularly great improvements in geometric consistency (MET3R: 0.3026 vs. 0.4522 for GEN3C). The superior performance confirms that our video diffusion model effectively follows the geometric guidance from the Cache, enabling precise trajectory control while maintaining photorealistic synthesis quality. Qualitative comparisons are shown in Fig.~\ref{fig:qaulitatively}.

\begin{table}[t]
\centering
\caption{\textbf{Quantitative comparison on sparse 360° views-to-video generation on Habitat~\cite{habitat19iccv} dataset.} $\downarrow$ indicates lower is better, $\uparrow$ indicates higher is better.}
\label{tab:sparse_view}
    \vspace{-3mm}
\resizebox{\columnwidth}{!}{
\begin{tabular}{lccccc}
\toprule
\textbf{Method} & \textbf{FVD} $\downarrow$ & \textbf{SSIM} $\uparrow$ & \textbf{PSNR} $\uparrow$ & \textbf{LPIPS} $\downarrow$ & \textbf{MET3R} $\downarrow$ \\
\midrule
ViewCrafter & 778.207 & 0.193 & 11.833 & 0.398 & 0.5061 \\
TrajectoryCrafter & 690.322 & 0.216 & 12.223 & 0.461 & 0.6741 \\
GEN3C & 511.039 & 0.481 & 17.307 & 0.195 & 0.4522 \\
\midrule
Pantheon360 (Ours) & \textbf{450.696} & \textbf{0.756} & \textbf{20.392} & \textbf{0.091} & \textbf{0.3026} \\
\bottomrule
\end{tabular}
}
\end{table}

\subsection{Two-View 360° Novel View Synthesis}

We further apply Pantheon360 to a challenging sparse-view novel view synthesis setting, where only two 360° views are provided and we generate novel views between them. This task is particularly relevant for synthesizing continuous videos from sparse Google Maps Street View panoramas.

\begin{figure}[bt!]
    \centering
    \includegraphics[width=1.0\linewidth]{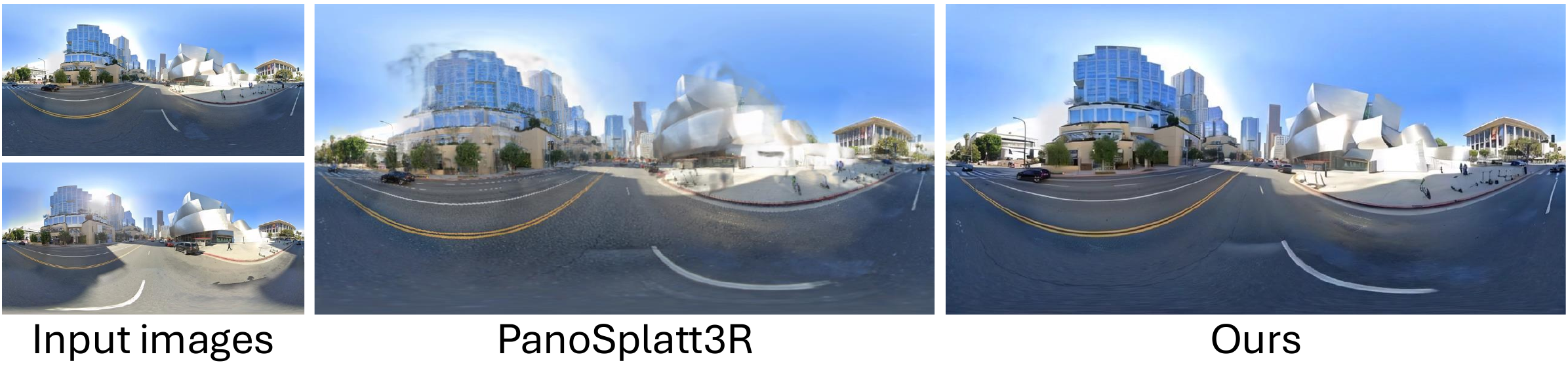}
    \vspace{-6mm}
    \caption{\textbf{Comparison to PanoSplatt3R~\cite{ren2025panosplatt3r}.} Our method produces geometrically accurate interpolations with clean structure, while PanoSplatt3R exhibits visible artifacts and geometric distortions.}
    \label{fig:panosplatt3r}
\end{figure}

\paragraph{Results.} 
As shown in Fig.~\ref{fig:panosplatt3r}, our method demonstrates superior geometric accuracy. While PanoSplatt3R produces geometrically inconsistent results with visible distortions, Pantheon360 maintains correct geometric structure throughout the synthesized trajectory.

\begin{figure*}[!t]
    \centering
    \vspace{-3mm}
    \includegraphics[width=1.0\linewidth]{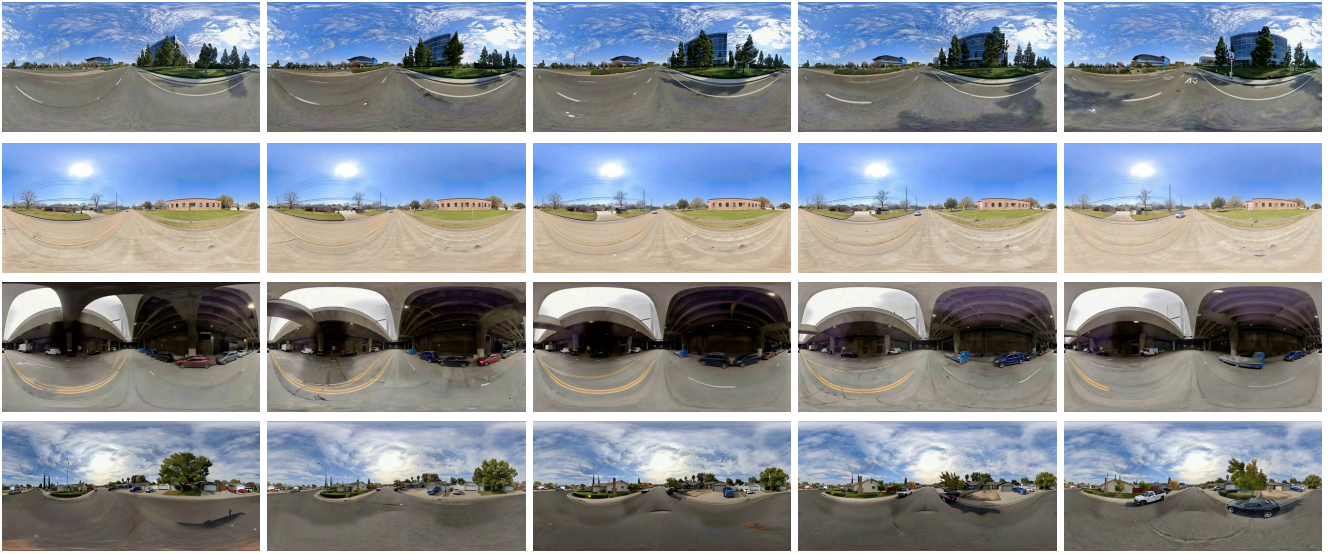}
    \vspace{-6mm}
    \caption{\textbf{Application.} Video Synthesis from Google Street View. Our method generates consistent 360° videos from sparse Google Street View imagery, enabling smooth navigation across extended trajectories.}
    \label{fig:google_street}
\end{figure*}

\begin{figure}[bt!]
    \centering
    \includegraphics[width=1.0\linewidth]{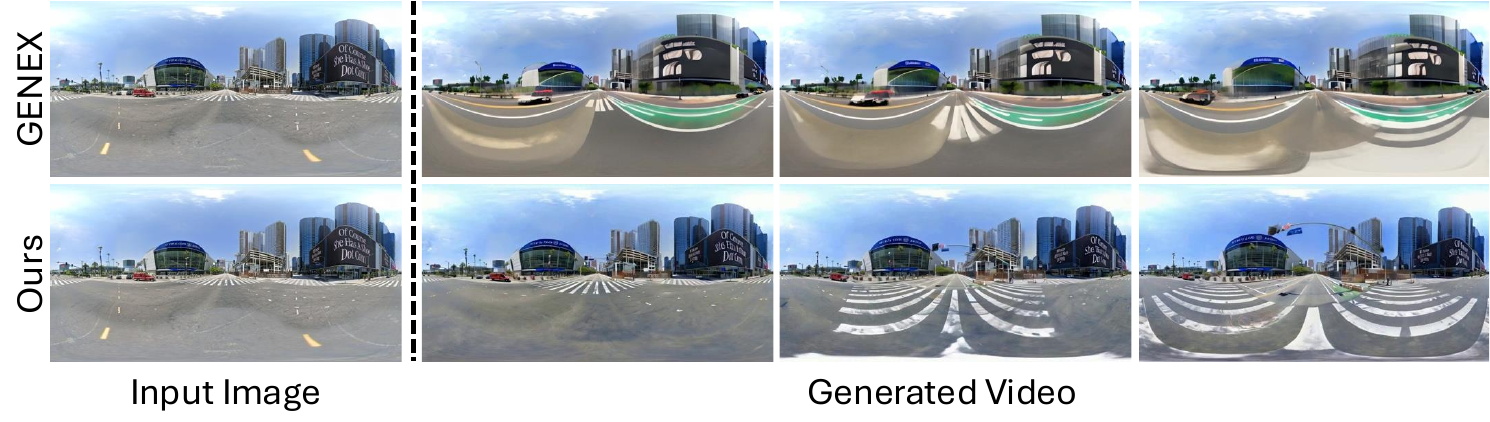}
    \vspace{-6mm}
    \caption{\textbf{Comparison to GenEX~\cite{lu2025genexgeneratingexplorableworld}.} Our method maintains consistent quality throughout the trajectory while GenEX's quality degrades rapidly with increasing geometric inconsistencies.}
    \label{fig:comp_genex}
\end{figure}

\subsection{Comparison with 360° World Models}

We compare Pantheon360 against GenEX~\cite{lu2025genexgeneratingexplorableworld}, a 360° world model designed for high-level action control. We evaluate both methods on Google Maps Street View panoramas with a simple forward motion trajectory.

\paragraph{Results.} 
As shown in Fig.~\ref{fig:comp_genex}, Pantheon360 maintains consistent quality and accurately follows the prescribed trajectory. In contrast, GenEX's quality degrades rapidly over frames with increasing geometric inconsistencies. Our explicit 3D Cache framework demonstrates superior temporal stability and geometric accuracy.

\subsection{Ablation Study}
\label{sec:ablation}

We evaluate four model variants to validate our dual-anchor latent fusion mechanism: (1) Single: conditioned only on the start frame, (2) Single+Latent Fusion: (1) with latent fusion, (3) Dual: conditioned on both start and end frames, and (4) Dual+Latent Fusion (our full method): (3) with latent fusion. We test on 30 Google Maps Street View scenes, measuring end frame alignment (PSNR, SSIM, LPIPS), short-term warping error (STWE), and interpolation error (IE).

\paragraph{Results.} 
As shown in Table~\ref{tab:ablation}, the Single model achieves the best temporal consistency but poor end frame alignment (20.92 PSNR). Dual anchor conditioning improves convergence (27.86 PSNR) while maintaining reasonable consistency. Our full method, Dual+Latent Fusion, achieves the best overall performance (28.95 PSNR, 7.44 IE), demonstrating that latent fusion effectively mitigates geometric inconsistencies while ensuring smooth interpolation.

\begin{table}[t]
\centering
\caption{\textbf{Ablation study on latent fusion for interpolation.} STWE refers to Short-Term Warping Error, and IE refers to Interpolation Error. $\downarrow$ indicates lower is better, $\uparrow$ indicates higher is better.}
\label{tab:ablation}
    \vspace{-3mm}
\resizebox{\columnwidth}{!}{
\begin{tabular}{lccccc}
\toprule
\textbf{Method} & \textbf{STWE} $\downarrow$ & \textbf{IE} $\downarrow$ & \textbf{PSNR} $\uparrow$ & \textbf{SSIM} $\uparrow$ & \textbf{LPIPS} $\downarrow$ \\
\midrule
Single & 0.124 & 4.784 & 20.921 & 0.661 & 0.271 \\
Single+Latent Fusion & 0.420 & 12.083 & 28.006 & 0.817 & 0.112 \\
Dual & 0.419 & 8.120 & 27.860 & 0.817 & 0.093 \\
\midrule
Dual+Latent Fusion (Ours) & 0.395 & 7.437 & \textbf{28.948} & \textbf{0.830} & \textbf{0.088} \\
\bottomrule
\end{tabular}
}
\end{table}

\begin{figure*}[!t]
    \centering
    \vspace{-3mm}
    \includegraphics[width=1.0\linewidth]{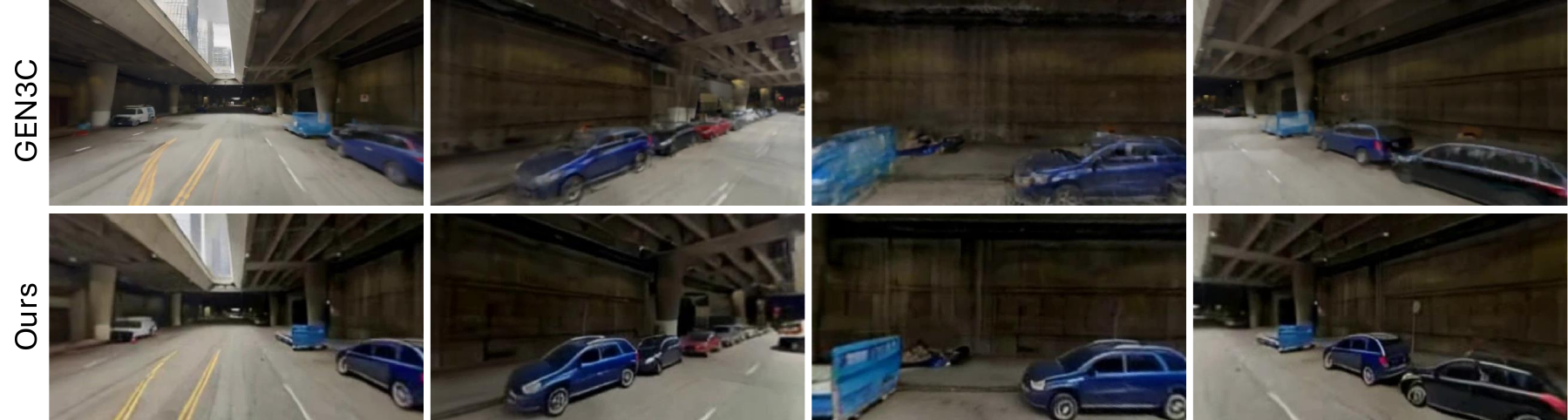}
    \vspace{-6mm}
    \caption{\textbf{Novel View Synthesis on Google Maps Street View.} Our method produces geometrically accurate renderings across different viewing angles with consistent structures. GEN3C~\cite{ren2025gen3c} suffers from ghosting artifacts, geometric distortions, and inter-view inconsistencies.}
    \label{fig:vis_googlemap}
\end{figure*}

\subsection{Applications}

\begin{figure}[!t]
    \centering
    \includegraphics[width=1.0\linewidth]{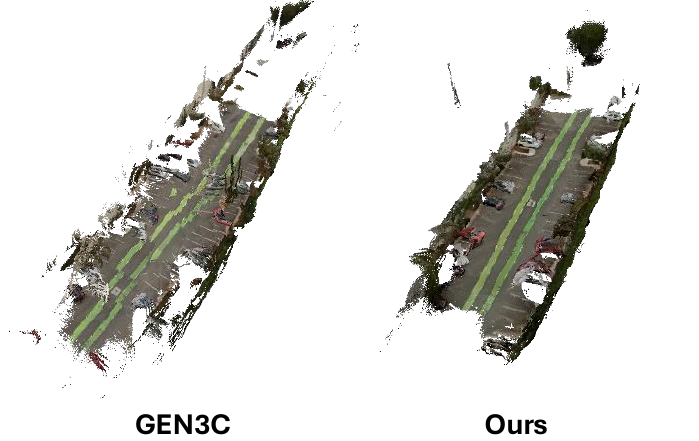}
    \vspace{-6mm}
    \caption{\textbf{3D Point Cloud Reconstruction Quality.} We reconstruct 3D point clouds from generated videos using $\pi^3$~\cite{wang2025pi3}. Our method yields dense, structurally coherent reconstructions (right), while GEN3C~\cite{ren2025gen3c} produces sparse, fragmented results (left), demonstrating our superior 3D consistency.}
    \label{fig:3d_cache}
\end{figure}

\paragraph{Video Synthesis from Sparse Street View Data.} 
We demonstrate Pantheon360 for synthesizing continuous navigation videos from sparse Google Maps Street View imagery. Our model's strong convergence to anchor frames enables sequential chaining: the final frame of one segment serves as the anchor for the next, allowing indefinite trajectory extension with global geometric consistency. As shown in Fig.~\ref{fig:vis_googlemap} and Fig.~\ref{fig:3d_cache}, our method produces geometrically accurate renderings with consistent object structures across different viewing angles. When reconstructing 3D point clouds from generated videos using PI3~\cite{wang2025pi3}, our method yields dense, structurally coherent reconstructions while GEN3C produces sparse, fragmented results, validating our superior geometric consistency. Fig.~\ref{fig:google_street} demonstrates smooth, coherent navigation videos across extended trajectories.

\paragraph{360° Video Stabilization.} 
We demonstrate video stabilization using synthetically perturbed Habitat trajectories~\cite{habitat19iccv}. Our pipeline extracts keyframes, reconstructs a 3D Cache, defines a smoothed trajectory $C_{smooth}$, and synthesizes stabilized video. By explicitly re-rendering scene geometry, Pantheon360 maintains temporal coherence and geometric consistency across the full 360° view. Video results are provided in supplementary materials.

\section{Conclusion}
\label{sec:conclusion}
We present Pantheon360, a framework for controllable 360° video generation with precise camera trajectory control through an explicit 3D Cache. By decoupling geometric reasoning from photorealistic synthesis, our approach generates temporally consistent videos with superior cross-view coherence. Experiments demonstrate state-of-the-art performance, and we showcase practical applications in Street View synthesis and video stabilization.

\paragraph{Limitations.} 
While our model can handle dynamic objects through learned motion priors, explicit control over object-level dynamics remains challenging. The 3D Cache primarily encodes static scene geometry, with dynamic motion relying on the diffusion model's learned priors. Future work could incorporate explicit motion representations to enable fine-grained control over object dynamics.

{\small
\bibliographystyle{ieeenat_fullname}
\bibliography{11_references}
}

\ifarxiv \clearpage \appendix \section{Overview}
\label{sec:overview}

This supplementary material provides additional details to support the main paper. Section~\ref{sec:implementation_details} provides comprehensive implementation details, including training and inference configurations, 3D Cache reconstruction settings, and dual-anchor latent fusion mechanism. Section~\ref{sec:data_curation} describes the detailed data curation pipeline for our training dataset, including quality filtering criteria and automatic trajectory annotation. Section~\ref{sec:additional_results} presents additional experimental results including a 3D Cache ablation study, runtime and memory analysis, 3D consistency validation via point cloud reconstruction, robustness analysis, and closed-loop trajectory validation.

In addition to this document, we provide an interactive HTML interface with supplementary videos demonstrating our method's capabilities on various tasks, including Google Street View synthesis, extended trajectory generation, and video stabilization. We also include qualitative comparisons with other state-of-the-art methods. 

\section{Implementation Details}
\label{sec:implementation_details}

\subsection{Training and Inference Details}

Our model is initialized from Argus~\cite{tan2025argus}, which is initialized from Stable Video Diffusion-I2V-XL model \cite{blattmann2023stable}. We train two separate models with identical configurations: (1) a single-anchor model conditioned on the first frame, and (2) a dual-anchor model conditioned on both start and end frames for interpolation tasks.

Both models are trained at $512 \times 1024$ resolution (height $\times$ width) in equirectangular format for 50,000 iterations, requiring approximately 5 days on 4 A100 GPUs. We set the sequence length $T = 25$ frames and train with a stronger noise schedule of $(P_{\text{mean}}, P_{\text{std}}) = (1, 1)$. We use the AdamW optimizer with a learning rate of $1 \times 10^{-5}$ and a batch size of 16. The training employs mixed precision (FP16) and gradient checkpointing for memory efficiency. At inference time, we use 25 denoising steps with a guidance scale of 5.0. For extended trajectory synthesis, we chain multiple generation segments by using the final frame of one segment as the anchor for the next.

\subsection{3D Cache Reconstruction}

We use $\pi^3$ (Pi3) \cite{wang2025pi3} as our primary 3D reconstruction foundation model. Since Pi3 is designed for perspective images, we first convert each 360° equirectangular input frame into multiple perspective views before feeding them into the model.

\noindent\textbf{Perspective View Extraction.}
For each 360° input frame, we use the Equi2Pers converter to extract perspective views with 90° horizontal field of view and output resolution of $768 \times 512$ (width $\times$ height). We sample at two pitch angles: 0° (horizontal view) and 60° (downward view toward the floor), excluding ceiling views to avoid sky regions. For yaw sampling, we use 50\% overlap between adjacent views, sampling at 45° intervals, resulting in 8 views per pitch level for complete 360° coverage. In total, we extract 16 perspective views per 360° frame (8 horizontal + 8 floor views).

These 16 perspective views are fed into PI3 to produce dense point cloud predictions with associated confidence scores. The predictions from all views are merged in a common coordinate frame defined by the camera poses estimated by Pi3.

\noindent\textbf{Point Cloud Filtering.}
We apply multiple filtering strategies to ensure high-quality 3D Cache. For confidence filtering, we apply a confidence threshold of 0.25, converting the raw confidence scores to probabilities via sigmoid function before thresholding. For edge filtering, we detect depth discontinuities using a relative tolerance of 0.03 and set their confidence to zero to remove unreliable edge points. Finally, we implement sky masking to remove unreliable sky points, which typically lack geometric structure and can introduce artifacts.

The filtered point clouds from all 16 views are merged to form the complete 3D Cache. Our framework is compatible with other 3D reconstruction methods such as VGGT \cite{wang2025vggt}, DUSt3R \cite{wang2024dust3r}, and MASt3R \cite{leroy2024mast3r}.

\subsection{Dual-Anchor Latent Fusion}

For our dual-anchor interpolation model, we employ the latent fusion technique from Time Reversal Fusion~\cite{feng2024explorative} to blend information from both start and end anchor frames. This approach is particularly effective when the 3D Cache quality is suboptimal due to sparse input views, where direct geometric conditioning alone may lead to discontinuities between the interpolated video and the target end frame.

\noindent\textbf{Bidirectional Geometric Conditioning.}
Given start and end frames, we reconstruct a 3D Cache from both anchor frames and render it along the target trajectory in two directions. The forward rendering produces a geometry video $V_{\text{geo}}^{\text{fwd}}$ by rendering from the start frame's viewpoint toward the end frame. The backward rendering produces $V_{\text{geo}}^{\text{bwd}}$ by rendering from the end frame's viewpoint toward the start frame (i.e., the trajectory is reversed). Both geometric videos are encoded into latent space as $v_{\text{fwd}} = \mathcal{E}(V_{\text{geo}}^{\text{fwd}})$ and $v_{\text{bwd}} = \mathcal{E}(V_{\text{geo}}^{\text{bwd}})$, where $\mathcal{E}$ denotes the VAE encoder.

\noindent\textbf{Latent Fusion Process.}
The fusion process operates in the latent space during the denoising procedure. At each denoising timestep $t$, we perform two separate denoising passes with different geometric and semantic conditioning. The forward pass is conditioned on the start frame features $c_s$ and forward geometric scaffold $v_{\text{fwd}}$, computing $\boldsymbol{x}_{t-1,s} = \Phi(\boldsymbol{x}_t, c_s, v_{\text{fwd}}, t)$. The backward pass is conditioned on the end frame features $c_e$ and backward geometric scaffold $v_{\text{bwd}}$, computing $\boldsymbol{x}_{t-1,e} = \Phi(\boldsymbol{x}_t, c_e, v_{\text{bwd}}, t)$, where $\Phi$ represents our denoising U-Net. 

These two predictions are then fused using a simple averaging strategy to produce the final denoised latent: $\boldsymbol{x}_{t-1} = \frac{1}{2}(\boldsymbol{x}_{t-1,s} + \boldsymbol{x}_{t-1,e})$. This fusion effectively combines the geometric information from both directions, helping to resolve inconsistencies and ensuring smooth convergence to the end frame while maintaining temporal coherence throughout the interpolated sequence.

Note that we do not employ the noise injection refinement proposed in~\cite{feng2024explorative} (setting $t_0 = 0$ and $M = 0$) to maintain faster inference speed while still achieving effective interpolation results as demonstrated in our ablation study (Table 3 in the main paper). The bidirectional geometric conditioning alone provides sufficient guidance for high-quality interpolation.

\section{Data Curation and Preparation}
\label{sec:data_curation}

\subsection{Quality Filtering Pipeline}

We build our training dataset starting from a curated subset prepared by \cite{wallingford2024image}, which selected approximately 100,000 high-quality video clips from the 360-1M dataset. We apply additional comprehensive filtering to further ensure training data quality suitable for our trajectory-controlled generation task.

\noindent\textbf{Format Validation.}
We first filter out mislabeled videos using two methods. For dual fisheye detection, we use Hough Circle detection to identify dual-fisheye format videos (two circular regions side-by-side), which are not true equirectangular panoramas. We sample 4 frames per video and reject videos where more than 90\% of frames contain dual circles. For perspective detection, we analyze boundary smoothness to detect perspective videos mislabeled as 360°. Videos with boundary smoothness greater than 0.25 are rejected.

\noindent\textbf{Motion Quality Assessment.}
Since our method requires meaningful camera motion, we filter videos with insufficient or problematic motion. For optical flow analysis, we compute optical flow using RAFT \cite{teed2020raft} at 1 FPS sampling rate with input resolution of $512 \times 256$. We apply equirectangular-aware weighting (latitude-based cosine weighting) to account for polar distortion. Videos with 75th percentile flow magnitude less than 3.0 pixels are rejected as static. For cut detection, we detect abrupt scene cuts using PySceneDetect to avoid training on concatenated clips. Videos with single-frame cut ratio greater than 0.3 or overall cut ratio greater than 0.2 are rejected.

\noindent\textbf{Content Quality Filtering.}
We remove low-quality or inappropriate content through two mechanisms. For image set detection, we detect slideshows by computing frame-to-frame MSE at 1 FPS. Videos with minimum MSE less than 1.0 (indicating identical consecutive frames) are rejected. For static region detection, we analyze top and bottom regions at multiple height ratios (from 1\% to 80\% of frame height) to detect static overlays or borders. For the 20\% height region, videos with MSE less than 1.0 in either top or bottom regions are rejected as they likely contain static UI elements or watermarks.

\noindent\textbf{Trajectory Annotation Filtering.}
After applying ViPE \cite{huang2025vipe} for automatic trajectory annotation, we additionally filter videos where ViPE fails to produce reliable results. This includes videos with insufficient camera baseline (too little camera motion for robust pose estimation), videos with too few SLAM feature points (indicating textureless or highly repetitive scenes), and videos where ViPE's optimization fails to converge.

\noindent\textbf{Filtering Statistics.}
Starting from the 100K curated subset, our additional filtering pipeline retains approximately 55,000 high-quality videos, corresponding to a retention rate of 55\%. This ensures that our final training dataset contains only videos with clear 360° format, sufficient motion, no static artifacts, and successful trajectory annotations. Each retained video is a 5-second clip, providing diverse camera trajectories and scene content for training.

\section{Additional Experimental Results}
\label{sec:additional_results}

\subsection{3D Cache Ablation Study}

We conduct an ablation study by progressively dropping points from the 3D Cache to quantify the importance of geometric conditioning $V_{\text{geo}}$. We report results on both Web360 (Table~\ref{tab:ablation_web360}) and Habitat (Table~\ref{tab:ablation_habitat}) benchmarks.

\begin{table}[h]
\centering
\small
\setlength{\tabcolsep}{3pt}
\begin{tabular}{l|ccccc}
\toprule
Drop Ratio & FVD$\downarrow$ & SSIM$\uparrow$ & PSNR$\uparrow$ & LPIPS$\downarrow$ & MET3R$\downarrow$ \\
\midrule
0\% (Ours) & \textbf{356.2} & \textbf{0.746} & \textbf{22.84} & \textbf{0.065} & \textbf{0.284} \\
25\% & 382.6 & 0.708 & 22.10 & 0.087 & 0.318 \\
50\% & 427.9 & 0.643 & 20.76 & 0.124 & 0.372 \\
75\% & 496.4 & 0.539 & 18.85 & 0.178 & 0.446 \\
100\% (w/o $V_{\text{geo}}$) & 553.3 & 0.421 & 16.93 & 0.251 & 0.523 \\
\bottomrule
\end{tabular}
\caption{\textbf{3D Cache ablation on Web360.} Performance degrades consistently as more points are dropped from the 3D Cache.}
\label{tab:ablation_web360}
\end{table}

\begin{table}[h]
\centering
\small
\setlength{\tabcolsep}{3pt}
\begin{tabular}{l|ccccc}
\toprule
Drop Ratio & FVD$\downarrow$ & SSIM$\uparrow$ & PSNR$\uparrow$ & LPIPS$\downarrow$ & MET3R$\downarrow$ \\
\midrule
0\% (Ours) & \textbf{450.7} & \textbf{0.756} & \textbf{20.39} & \textbf{0.091} & \textbf{0.303} \\
25\% & 489.2 & 0.712 & 19.55 & 0.118 & 0.349 \\
50\% & 551.8 & 0.638 & 18.12 & 0.162 & 0.414 \\
75\% & 637.5 & 0.524 & 16.28 & 0.231 & 0.502 \\
100\% (w/o $V_{\text{geo}}$) & 724.2 & 0.387 & 14.22 & 0.319 & 0.597 \\
\bottomrule
\end{tabular}
\caption{\textbf{3D Cache ablation on Habitat.} Consistent with Web360, removing $V_{\text{geo}}$ leads to significant performance degradation across all metrics.}
\label{tab:ablation_habitat}
\end{table}

Without $V_{\text{geo}}$, the model reduces to standard image-to-video generation. While it can still produce visually plausible results, geometric correctness cannot be fully guaranteed. Moreover, since we explicitly rely on the rendered point cloud as a condition for camera control, removing $V_{\text{geo}}$ makes precise trajectory control infeasible.

\subsection{Runtime and Memory Analysis}

We provide detailed runtime and memory analysis on Google Map data using a single A100 GPU at $1024\times512$ resolution. As shown in Table~\ref{tab:runtime}, diffusion denoising is the main bottleneck, accounting for approximately 80\% of total inference time. Our method runs entirely on a single GPU, and the modular framework is compatible with faster diffusion models, which can reduce inference time in future work.

\begin{table}[h]
\centering
\scriptsize
\setlength{\tabcolsep}{2pt}
\begin{tabular}{l|cc|ccc|c|c}
\toprule
 & Input & Output & \multicolumn{3}{c|}{Time (s)} & & \\
Setting & Views & Frames & Recon. & Render & Diff. & Total & Mem. \\
\midrule
Single-view & 1 & 25 & 34s & 2s & 163s & 199s & 30GB \\
Interp. (w/ latent fusion) & 2 & 25 & 50s & 5s & 320s & 375s & 41GB \\
Long traj. (w/ latent fusion) & 5 & 100 & 74s & 7s & 1284s & 1365s & 41GB \\
\bottomrule
\end{tabular}
\caption{\textbf{Runtime and memory analysis} across different settings on a single A100 GPU. Recon., Render, and Diff. denote the time for 3D point cloud reconstruction, geometry rendering, and diffusion denoising, respectively.}
\label{tab:runtime}
\end{table}

\subsection{3D Consistency Validation via Point Cloud Reconstruction}

To validate that our generated videos maintain 3D geometric consistency while successfully hallucinating unseen regions, we reconstruct 3D point clouds from both the reference image and our generated video using Pi3~\cite{wang2025pi3}. As shown in Figure~\ref{fig:pcd_reconstruction}, the point cloud reconstructed from the reference image (Before) contains only the visible geometry from the input viewpoint. In contrast, the point cloud reconstructed from our generated video (After) is significantly more complete, successfully hallucinating previously occluded regions while maintaining consistency with the original scene structure. This demonstrates that our method not only preserves 3D geometric consistency but also generates plausible geometry for unseen areas, resulting in a more complete 3D reconstruction. This capability is essential for applications requiring comprehensive scene understanding from limited input views.

\begin{figure}[t]
    \centering
    \includegraphics[width=0.8\linewidth]{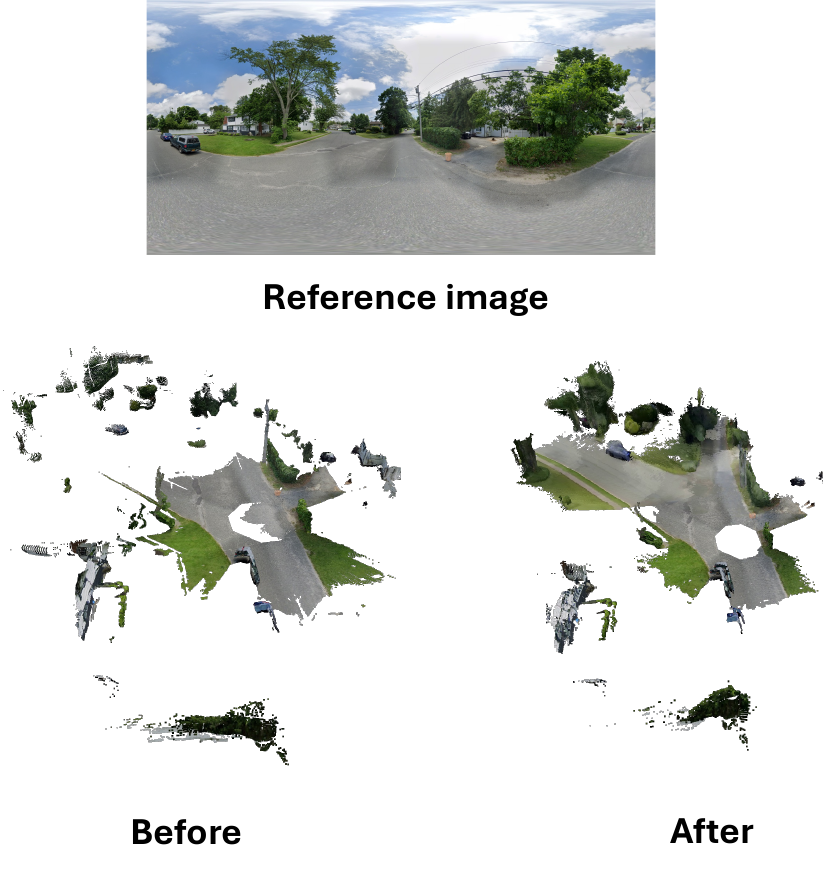}
    \caption{\textbf{3D Point Cloud Reconstruction Quality.} We reconstruct 3D point clouds using Pi3~\cite{wang2025pi3} from the reference image (Before) and our generated video (After). Our generated video produces a more complete 3D reconstruction by successfully hallucinating occluded regions while maintaining geometric consistency with the original scene.}
    \label{fig:pcd_reconstruction}
\end{figure}

\subsection{Robustness Analysis and Failure Cases}

Our method is robust to moderate reconstruction errors due to the video diffusion prior. Even when the rendered 3D Cache contains holes or inaccuracies (e.g., from dynamic objects or low-light conditions), our model can inpaint and refine these regions, as shown in Figure~\ref{fig:robustness}.

\begin{figure}[h]
    \centering
    \includegraphics[width=1.0\linewidth]{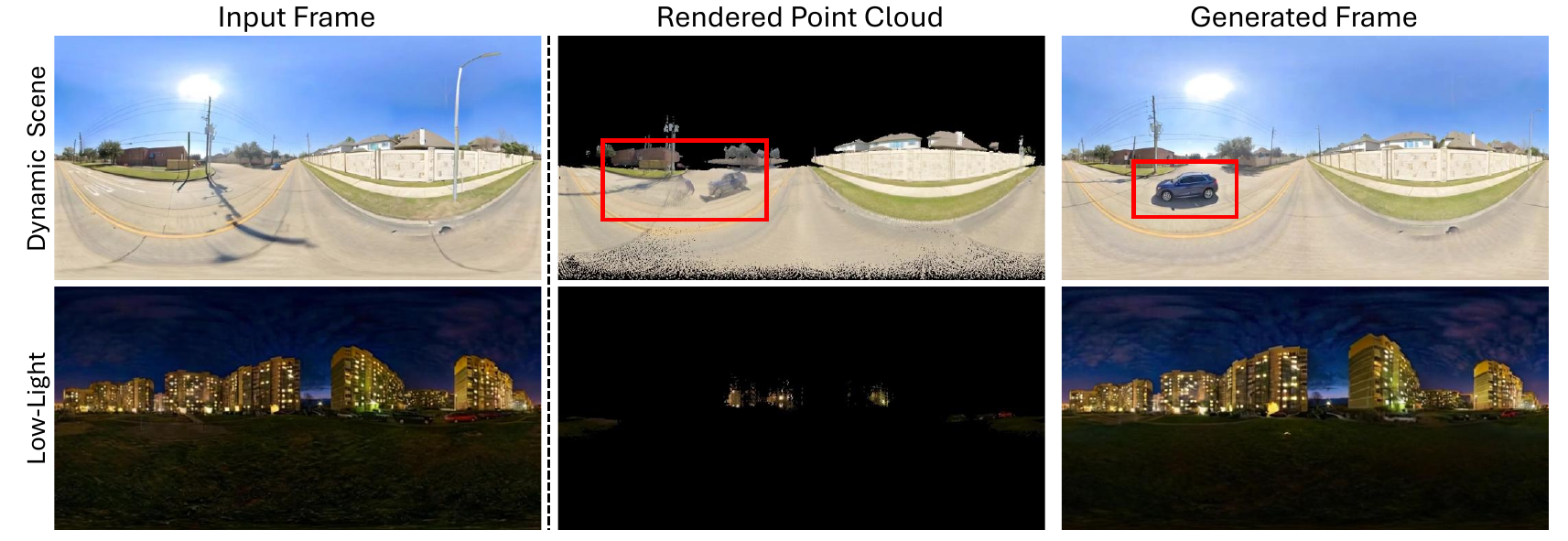}
    \caption{\textbf{Robustness to 3D Cache imperfections.} Our video diffusion prior can inpaint and refine regions where the 3D Cache contains holes or inaccuracies caused by dynamic objects or low-light conditions.}
    \label{fig:robustness}
\end{figure}

However, we identify two failure cases illustrated in Figure~\ref{fig:failure}: (1) crowded dynamic scenes with many moving objects, resulting in motion blur artifacts; (2) input 360° images with stitching artifacts that propagate through the pipeline. Addressing these limitations requires future advances in 4D reconstruction or higher-quality input capture.

\begin{figure}[h]
    \centering
    \includegraphics[width=1.0\linewidth]{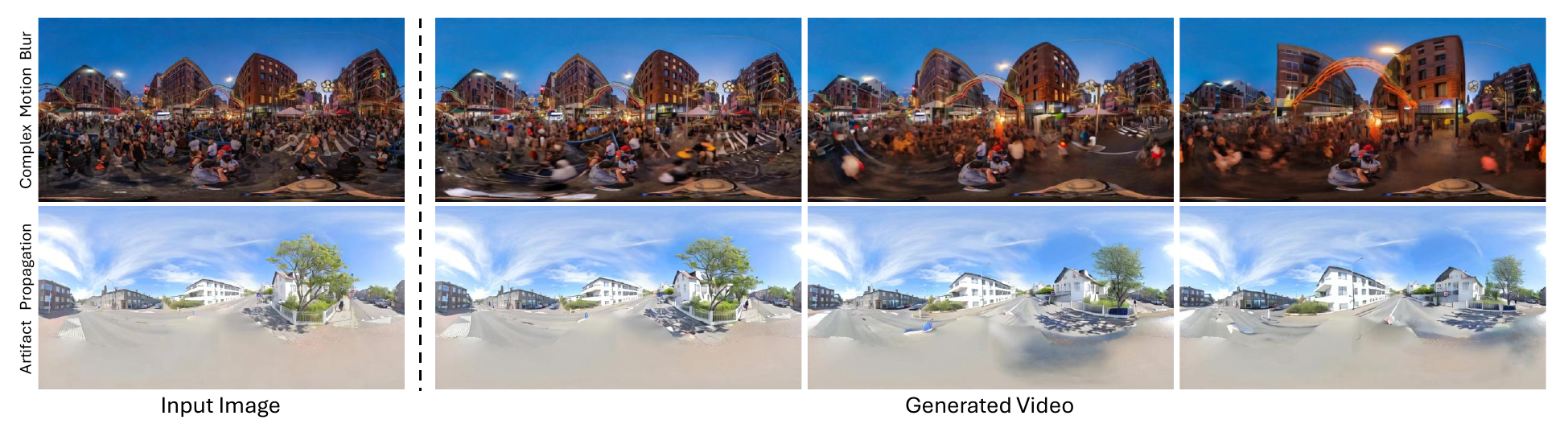}
    \caption{\textbf{Failure cases.} Our method struggles with (1) crowded dynamic scenes with many moving objects, and (2) input 360° images with stitching artifacts.}
    \label{fig:failure}
\end{figure}

\subsection{Closed-Loop Trajectory Validation}

We validate trajectory robustness on a closed-loop trajectory from Google Map data. Using sparse 360° panoramas as input, our method generates temporally consistent video when revisiting the starting region, as shown in Figure~\ref{fig:loop}. This is enabled by our 3D Cache, which provides persistent geometric grounding throughout the trajectory.

\begin{figure}[h]
    \centering
    \includegraphics[width=1.0\linewidth]{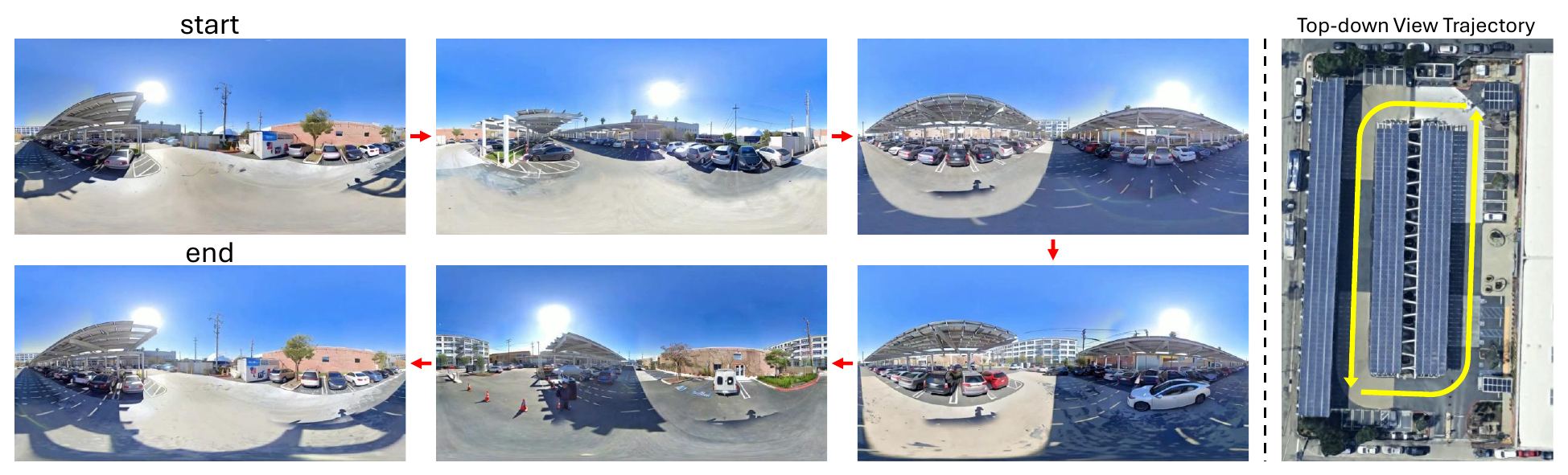}
    \caption{\textbf{Closed-loop trajectory validation.} Our method generates consistent video along a closed-loop trajectory, successfully revisiting the starting region without temporal inconsistencies.}
    \label{fig:loop}
\end{figure} \fi

\end{document}